\documentclass[10pt,twocolumn,letterpaper]{article}

\usepackage[pagenumbers]{cvpr} % To force page numbers, e.g. for an arXiv version
\usepackage{comment}

\definecolor{cvprblue}{rgb}{0.21,0.49,0.74}
\usepackage[pagebackref,breaklinks,colorlinks,allcolors=cvprblue]{hyperref}
\usepackage{graphicx}
\usepackage{colortbl}
\usepackage{amsmath}
\usepackage{multirow}
\usepackage{multicol}
\usepackage{amsfonts}       % blackboard math symbols
\usepackage{nicefrac}       % compact symbols for 1/2, etc.
\usepackage{microtype}      % microtypography
\usepackage{xcolor}         % colors
\usepackage{arydshln}
\definecolor{mygray}{RGB}{230,230,230}

\def\paperID{7151} % *** Enter the Paper ID here
\def\confName{CVPR}
\def\confYear{2026}

\title{Learning to See through Illumination Extremes with Event Streaming in\\ Multimodal Large Language Models}

\author{Baoheng Zhang$^{*}$~~ Jiahui Liu$^{*\dagger}$~~Gui Zhao~~ Weizhou Zhang~~ Yixuan Ma~~ Jun Jiang~~\\ Yingxian Chen~~ Wilton W.T.~Fok~~ Xiaojuan Qi$^{\ddagger}$~~Hayden Kwok-Hay So$^{\ddagger}$\\
The University of Hong Kong\\
\hspace{-12pt}\texttt{\small \{bhzhang, liujh, xjqi, hso\}@eee.hku.hk}\\
$^{*}$co-first authors~~~$^{\dagger}$project leader~~~$^{\ddagger}$corresponding authors
}

\begin{document}
\maketitle
\begin{abstract}
Multimodal Large Language Models (MLLMs) perform strong vision–language reasoning under standard conditions but fail in extreme illumination, where RGB inputs lose irrevocable structure and semantics. 
We propose Event-MLLM, an event-enhanced model that performs all-light visual reasoning by dynamically fusing event streams with RGB frames. Two key components drive our approach: an Illumination Indicator -- a learnable signal derived from a DINOv2 branch that represents exposure degradation and adaptively modulates event-RGB fusion -- and an Illumination Correction Loss that aligns fused features with non-degraded (normal-light) semantics in the latent space, compensating for information lost in extreme lighting.
We curate the first multi-illumination event-instruction corpus for MLLMs, with 2,241 event-RGB samples (around 6 QA pairs each) across diverse scenes and 17 brightness rates (0.05× – 20×), plus an instruct-following benchmark for reasoning, counting, and fine-grained recognition under extreme lighting. 
Experiments show that Event-MLLM markedly outperforms general-purpose, illumination-adaptive, and event-only baselines, setting a new state of the art in robust multimodal perception and reasoning under challenging illumination.
\end{abstract}    
\section{Introduction}
\label{sec:intro}

Multimodal Large Language Models (MLLMs) have recently demonstrated impressive advances in visual understanding, reasoning, and instruction following~\cite{alayrac2022flamingo,li2023blip2, liu2023llava, zhu2023minigpt4}. By coupling powerful vision encoders with large-scale language models, they can perform a broad spectrum of tasks -- from open-ended question answering to fine-grained visual reasoning~\cite{chen2024internvl, lu2024deepseekvl, yang2024llama3v, wang2024qwen2vl}. However, most existing MLLMs implicitly assume ideal visual conditions. Under extreme illumination, such as overexposure or near darkness, RGB images suffer irreversible structural and semantic information loss, leading to significant degradation in model perception and reasoning. The inability to perceive key scene details not only limits downstream understanding but also results in hallucinations and factual errors, revealing a critical blind spot in current MLLMs’ robustness.

\begin{figure}[t]
\centering
\vspace{0.5cm}
\includegraphics[width=1.0\linewidth]{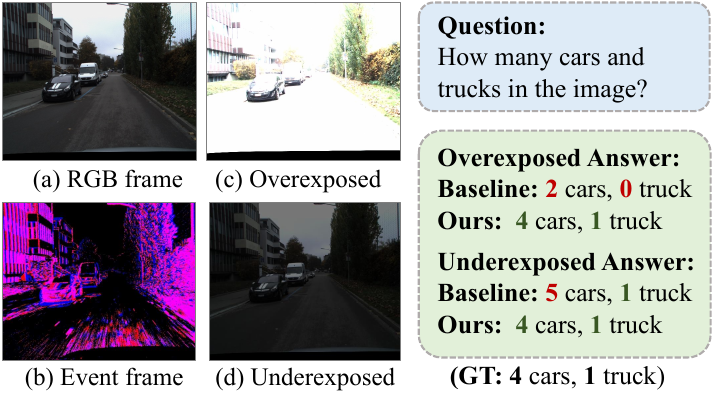}
\vspace{-0.7cm}
\caption{Extreme lighting scenarios are demonstrated. When the normal (a) RGB frame can not be obtained, only (c) overexposed or (d) underexposed frame can be used as input. Existing baselines often produce illusions due to information degradation, while our method cleverly incorporates information from the (b) event frame, enabling the model to provide more accurate responses.}
\vspace{-0.6cm}
\label{fig:intro}
\end{figure}

Several strategies have been proposed to mitigate the degradation induced by illumination. Image restoration pipelines attempt to pre-enhance visual inputs before MLLM inference~\cite{cheng2024multimodal}, yet these methods often introduce artificial textures or distort semantics. Another mainstream direction treats the MLLM as an intelligent controller that calls external image-enhancement tools~\cite{lin2025jarvisir,wang2025mllmtoolmultimodallargelanguage,jiang2025multiagentimagerestoration}, improving perception indirectly but without enhancing intrinsic model robustness. Other works~\cite{zhang2024qinstruct} rely on specialized low-light encoders or meta-quality assessment modules to detect exposure issues, but fail to guarantee consistent content-level understanding~\cite{wu2024adamerger}. Overall, existing approaches are reactive -- they aim to recover degraded information rather than proactively preventing information loss within the MLLM reasoning process.

Event cameras offer a compelling alternative. Operating asynchronously with microsecond temporal resolution and a dynamic range exceeding 120 dB, they capture brightness changes rather than absolute intensity~\cite{lichtsteiner2008}. This property enables event streams to retain rich structural cues even when RGB sensors saturate or fail. Earlier research has shown that fusing event data with RGB images can effectively restore details in low-level vision tasks such as denoising, deblurring, and HDR reconstruction~\cite{lin2022coherent, liang2024towards, jin2024towards,jiang2024evlight++}. More recent studies have extended this paradigm to detection and segmentation~\cite{ram2022enhancing}. However, these event-based methods remain confined to perception-level understanding and have not been integrated into MLLMs for reasoning or instruct-following tasks.

Despite their complementary potential, leveraging event data within MLLMs introduces new challenges. (a) \textit{Modality Alignment}: event representations differ fundamentally from frame-based RGB features, complicating cross-modal fusion and alignment in the language space. (b) \textit{Adaptive Fusion}: illumination degradation varies spatially and temporally, requiring dynamic weighting between event and RGB streams. (c) \textit{MLLM Fine-tuning}: in the absence of clean references, it remains unclear how to guide the model to fuse modalities toward information-rich semantics rather than redundant correlations. These challenges highlight the need for a principled mechanism that allows MLLMs to learn when and how to rely on event information under varying illumination.

We present an event-enhanced multimodal large language model, \textbf{Event-MLLM}, that dynamically fuses event streams with RGB frames for robust, all-light visual reasoning. Our approach hinges on two innovations: (1) \textit{Illumination Indicator} -- a learnable signal from a DINOv2 branch that quantifies exposure degradation and adaptively modulates the event-RGB fusion; and (2) \textit{Illumination Correction Loss} -- a feature-space objective that aligns fused representations to non-degraded (normal-light) semantics, effectively distilling structure lost in extreme lighting. These components enable illumination-aware reasoning during inference using only degraded RGB and the paired event input. Additionally, we introduce the first large-scale, multi-illumination event-instruction dataset, covering diverse indoor/outdoor scenes across brightness rates from 0.05× to 20×, and a comprehensive instruction-following QA benchmark that evaluates reasoning, localization, and fine-grained recognition under illumination extremes.

Extensive experiments demonstrate that our designed model achieves state-of-the-art performance, significantly outperforming existing methods under different lighting conditions (as illustrated in Figure~\ref{fig:intro}), and setting a new standard for robust multimodal perception under challenging illumination. 
Our main contributions include:
\begin{itemize}
\item Event-Enhanced MLLM: We propose Event-MLLM, the first event-enhanced MLLM that fuses event frames with RGB frames to enable robust, all-light visual reasoning.
\item Illumination-Aware Fusion with Semantic Alignment: We introduce a learnable \textit{Illumination Indicator} (from a DINOv2 branch) that represents exposure degradation to adaptively modulate event–RGB fusion, together with an \textit{Illumination Correction Loss} that aligns fused features to normal-light semantics, distilling structure lost under extreme lighting.
\item Dataset \& Benchmark with SOTA Results: We curate the first multi-illumination event-instruction dataset spanning indoor/outdoor scenes across 0.05×–20× brightness and build a comprehensive instruct-following benchmark (reasoning, counting, and fine-grained recognition). Extensive experiments show consistent state-of-the-art performance over general MLLMs, event-only, and illumination-adaptive baselines.
\end{itemize}

\section{Related Works}
\label{sec:related}

\begin{figure*}[t]
\centering
\includegraphics[width=1.0\linewidth]{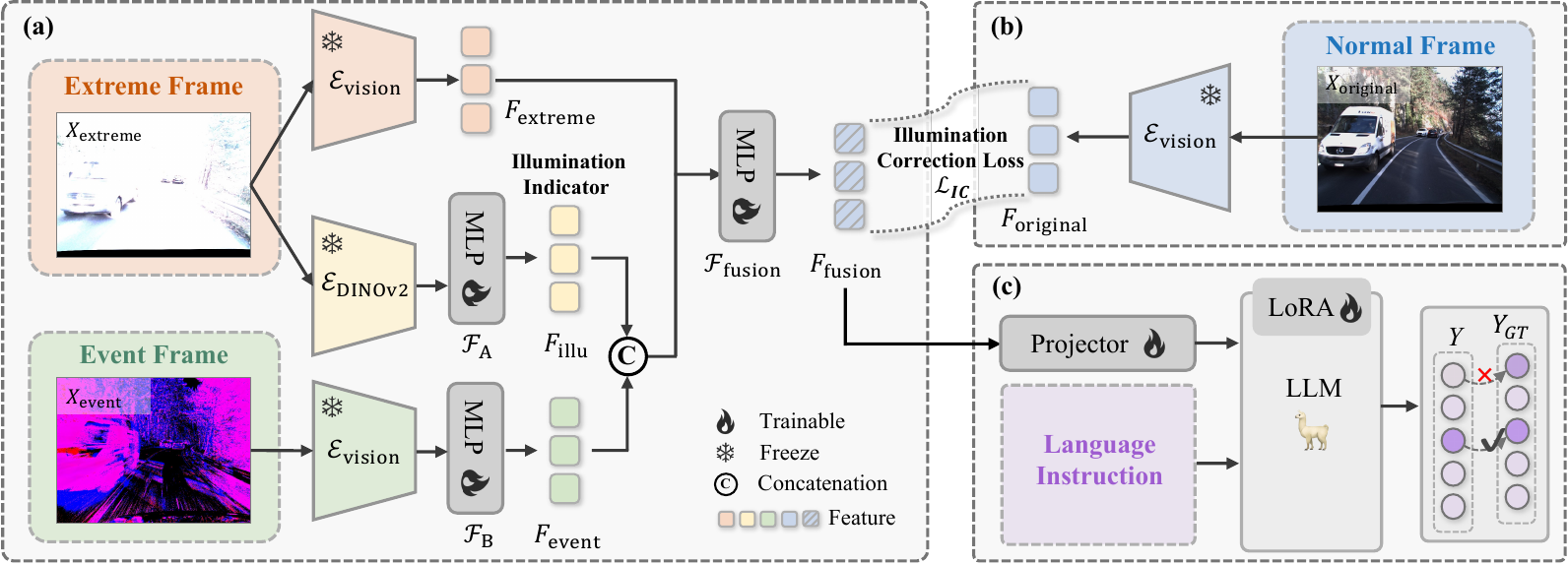}
\vspace{-0.5cm}
\caption{An overview of our proposed model is illustrated: (a) Features are extracted from extreme-light RGB frames by the original vision encoder and the DINOv2 vision encoder, respectively. These features are then fused with the features extracted from the event frame using our designed MLPs with our proposed Illumination Indicator ($F_{\text{illu}}$) to obtain fused features $F_{\text{fusion}}$ (Section~\ref{sec:met-fea}). (b) Features are extracted from normal-light RGB frames and used for representation learning with the fused features using our proposed Illumination Correction Loss $\mathcal{L}_{\mathrm{IC}}$ (Section~\ref{sec:met-ill}). (c) During training, the fused features participate in the fine-tuning of the entire MLLM, while (a) and (c) participate in inference (Section~\ref{sec:met-train}).}
\vspace{-0.5cm}
\label{fig:framework}
\end{figure*}

\subsection{Event based Vision Model}
Event based vision model has attracted significant research attention due to its unique characteristics, particularly high dynamic range and microsecond-level temporal resolution that surpass conventional imaging modalities~\cite{gao2024composable,zhang2024co}. Current research evolves along three main trajectories: 1) adapting event data to traditional vision tasks like classification, detection, segmentation, and depth estimation~\cite{ram2022enhancing,jiang2024evlight++, Gehrig2023RVT, GET_cls}; 2) fusing RGB and event modalities for low-level vision tasks including image enhancement, denoising, and deblurring~\cite{lin2022coherent,liang2024towards,jin2024towards}; and 3) developing event based foundation models for vision-language tasks~\cite{wu2023eventclip,liu2024eventgpt,li2025eventvl}.

Among these directions, leveraging the unique advantages of event data in extreme illumination conditions has emerged as a growing research focus, with recent works exploring RGB-event integration to enhance visual performance under challenging lighting, which evolves along two main streams: 1) establishing of aligned RGB-event datasets for various scenarios for lowlight recognition and enhancement ~\cite{jiang2024evlight++,yang2024hue, jin2024towards, li2024event}, including indoor and outdoor scenarios. 2) development of fusion methods that augment existing RGB-based frameworks with event data for improved performance under extreme illumination~\cite{liu2022lowlightvideoenhancementsynthetic, event_illu, liang2024robusteventguidedlowlightimage}.

However, leveraging event data to enhance Vision-Language Models (VLMs) remains unexplored. Previous work EventGPT~\cite{liu2024eventgpt} and EventVL~\cite{li2025eventvl} established the first end-to-end frameworks incorporating dedicated spatiotemporal modules for event stream understanding without RGB image fusing. Subsequent works like LLaFEA~\cite{zhou2025llafea} enhanced fine-grained reasoning through event-video fusion, while Talk-to-the-Events~\cite{xiong2025talktotheevents} explored multi-round dialogue capabilities. Nevertheless, these approaches overlook robustness across varying illumination conditions including both under-exposed and over-exposed.

\subsection{Vision Language Model}
The remarkable success of Large Language Models (LLMs) has catalyzed their integration into multimodal research~\cite{liu2024can,chen2025aligning,wu2024saco,xia2025dreamomni2multimodalinstructionbasedediting}, primarily focusing on establishing robust vision-language alignment through pioneering works like BLIP and Flamingo \cite{li2022blip,alayrac2022flamingo,li2023blip2,lu2019vilbert,tan2019lxmert}, as extensively surveyed in recent literature \cite{caffagni2024revolution,lin2024vila,yin2024survey,zhang2024mmllms,liu2024mmbench}. The paradigm shifted significantly with LLaVA's visual instruction tuning, which provided a public, data-efficient framework that sparked the open-source MLLM boom, yielding conversational agents including MiniGPT-4, InternVL, and the Llama-V series \cite{liu2023llava,zhu2023minigpt4,chen2024internvl,wang2024qwen2vl,lu2024deepseekvl,yang2024llama3v}. However, the image quality affects the model performance. The prevalent cascaded architecture—feeding visual encoder outputs to LLMs—reveals that input images constitute the performance bottleneck \cite{liu2023mmsafetybench,cui2024robustness,li2023pope,yang2023dawn}, where visual degradations like extreme brightness impair MLLM performance \cite{openai2023gpt4v}.

Among these directions, there are works that focus on addressing the extreme illumination in Vision language model. Recent robustness benchmarks\cite{liu2023mmsafetybench,zhang2024robustvlm,bai2025hmllm,schmitt2024challenger} quantitatively validated that leading MLLMs experience precipitous performance degradation under extreme brightness conditions. Preliminary works conduct different strategies to solve the illumination robustness problem, like creating specialized instruction datasets (Q-Instruct~\cite{zhang2024qinstruct}), restoring degraded inputs prior to understanding~\cite{wu2024adamerger}, or leveraging complementary modalities like depth~\cite{cheng2024multimodal}. However, these approaches overlook event camera, which demonstrates reliability in extreme illumination scenarios such as low-light and overexposed conditions where conventional RGB camera fails.

\section{Method}
\label{sec:method}

\subsection{Overview}
\label{sec:over}

\noindent \textbf{Task.} Our event-enhanced all-light cognitive MLLM aims to accurately comprehend visual scenes and follow language instructions under all-light illumination conditions. 
We train the model with an instruct-following dataset built upon visual triplet: normal RGB frames, extreme-light frames under different brightness rates, and event frames. 
During inference, the model receives only the degraded (extreme-light) frames and the corresponding event frames for all-light reasoning. 
The core objective is to learn how to adaptively fuse the two visual modalities, create a unified visual representation that is robust to illumination changes, and align fused features well with language space.

\noindent \textbf{Pipeline.} 
As illustrated in Figure~\ref{fig:framework}, our event-enhanced MLLM architecture processes dual visual input streams. 
Given a degraded (extreme-light) frame $X_{\text{extreme}}$ and a corresponding event frame $X_{\text{event}}$ as input, the original visual encoder $\mathcal{E}_{\text{vision}}$ extracts high-level semantic features, denoted as $F_{\text{extreme}}$ and $F_{\text{event}}$, respectively. 
A DINOv2~\cite{oquab2023dinov2} encoder $\mathcal{E}_{\text{DINOv2}}$ processes $X_{\text{extreme}}$ to capture illumination characteristics, yielding an illumination-aware feature $F_{\text{illu}}$. 
At the same time, the visual feature $F_{\text{original}}$ is extracted by $\mathcal{E}_{\text{vision}}$ with a normal-light original image $X_{\text{original}}$ as input.
During the training phase, these three types of features are fed into our proposed MLP modules, which generate a unified visual representation $F_{\text{fusion}}$. 
This representation is supervised by the feature $F_{\text{original}}$, so that the blended features can approximate the representation of the original image information closely. 
The fused feature $F_{\text{fusion}}$ is then mapped by a projection layer into the LLM's language space and fed into the LLM for end-to-end instruction-following fine-tuning. 
For inference, the model directly utilizes the adaptively fused features $F_{\text{fusion}}$ for reasoning with only $X_{\text{extreme}}$ and $X_{\text{event}}$ as input, without any $X_{\text{original}}$.

\noindent \textbf{Key Design.} 
Our key design consists: 
a feature fusion mechanism guided by a dynamic Illumination Indicator, which adaptively fuses the RGB and event modalities (see Section~\ref{sec:met-fea}); 
and an Illumination Correction Loss, serving as an implicit supervisory signal, aligns the fused visual representations toward the original visual representations within the language feature space (see Section~\ref{sec:met-ill}). 
In addition, we develop a specialized two-stage training strategy to optimize the MLLM effectively, as detailed in Section~\ref{sec:met-train}.

\subsection{Feature Fusion with Illumination Indicator}
\label{sec:met-fea}
To achieve robust visual understanding across all illumination conditions, we propose a fusion module that adaptively fuses event and extreme RGB features guided by a learnable \textbf{Illumination Indicator} (see Figure~\ref{fig:framework}). 
Rather than relying on hand-crafted fixed weighting for the fusion process, we use a learnable signal from the DINOv2 branch as an Illumination Indicator, which directly modulates how event features need to be fused into extreme features.
As shown in Figure~\ref{fig:framework}(a), we first encode the extreme feature $F_{\text{extreme}}$ 
by applying a frozen $\mathcal{E}_{\text{vision}}$ on $X_{\text{extreme}}$, which captures high-level semantics under extreme lightness condition:
\begin{equation}
    F_{\text{extreme}} = \mathcal{E}_{\text{vision}}(X_{\text{extreme}}).
\end{equation}

Meanwhile, a lightness feature $F_{\text{illu}}$ is derived by processing $X_{\text{extreme}}$ through a frozen DINOv2 encoder~\cite{oquab2023dinov2} followed by a learnable lightweight MLP, $\mathcal{F}_{\text{A}}$, which serves as our Illumination Indicator, encoding the global illumination degradation pattern:
\begin{equation}
    F_{\text{illu}} = \mathcal{F}_{\text{A}}\left(\mathcal{E}_{\text{DINOv2}}(X_{\text{extreme}})\right).
\end{equation}
Similarly, the event feature $F_{\text{event}}$ is extracted by passing $X_{\text{event}}$ through the $\mathcal{E}_{\text{vision}}$ and another learnable MLP, $\mathcal{F}_{\text{B}}$:
\begin{equation}
    F_{\text{event}} = \mathcal{F}_{\text{B}}\left(\mathcal{E}_{\text{vision}}(X_{\text{event}})\right).
\end{equation}

Based on these features, we perform the illumination-guided fusion. The lightness feature $F_{\text{illu}}$ and the event feature $F_{\text{event}}$ are first concatenated,
This step connects the robust event information with the specific lighting conditions, enabling the model to adaptively extract complementary information from events to enrich the degraded extreme RGB features under certain adverse illumination signaled by the Illumination Indicator.
Finally, the combined representation is concatenated with the extreme feature $F_{\text{extreme}}$ from the RGB frame. The resulting comprehensive feature is fed into a final fusion MLP $\mathcal{F}_{\text{fusion}}$, to produce:
\begin{equation}
    F_{\text{fusion}} = \mathcal{F}_{\text{fusion}} \left(  F_{\text{extreme}}, \text{Concat}[F_{\text{illu}}, F_{\text{event}}] \right),
\end{equation}
which is a refined visual representation that has been adaptively corrected for illumination degradation and is used for subsequent projection into the language space for multimodal all-light reasoning.

\begin{table*}
\caption{Compared to existing MLLMs of different types, our method significantly outperforms all other methods to achieve state-of-the-art performance on both multi-choice and object-counting tasks on all metrics. Furthermore, we take experiments with different sizes of Qwen~\cite{wang2024qwen2vl} as baselines and apply our design, where consistently achieves performance improvements (Best results are shown in \textbf{bold}).}
\vspace{-0.6cm}
\centering
\renewcommand{\arraystretch}{1.0}
\setlength{\tabcolsep}{8pt}
\begin{center}
\resizebox{1.0\linewidth}{!}{
\begin{tabular}{llcccc}
\toprule
\multirow{2}{*}{\textbf{MLLM Type}} & \multirow{2}{*}{\textbf{Method}} & \multicolumn{2}{c}{\textbf{Multi-Choice}} &  \multicolumn{2}{c}{\textbf{Object-Counting}}  \\
 & & \textbf{Accuracy (\%) $\uparrow$} & \textbf{F1 Score (\%) $\uparrow$} & \textbf{Accuracy (\%) $\uparrow$} & \textbf{MAE $\downarrow$} \\
\midrule
\multirow{3}{*}{General Model}& LLaVA-7B~\cite{liu2023llava} & 18.43 & 65.39 & 67.84 & 0.9957 \\
& Deepseek-VL~\cite{wu2024deepseekvl2mixtureofexpertsvisionlanguagemodels} & 7.00 & 55.70 & 54.87 & 1.0544\\
& InternVL~\cite{zhu2025internvl3exploringadvancedtraining} & 31.50 & 75.46 & 72.56 & 0.8641\\
\midrule
Event-specialized Model & EventGPT~\cite{liu2024eventgpt} & 3.16 & 73.60 & 67.41 & 1.0015 \\
\midrule
Illumination-aware Model & Q-Instruct~\cite{zhang2024qinstruct} & 15.86 & 65.09 & 66.87 & 0.9803 \\
\midrule
\multirow{4}{*}{Event-Enhanced Model} & \emph{Baseline-3B} & 29.86 & 77.04 & 69.66 & 0.6655 \\
& \cellcolor{mygray}\emph{Ours-3B} & \cellcolor{mygray}34.48 & \cellcolor{mygray}80.35 & \cellcolor{mygray}71.08 & \cellcolor{mygray}0.6127 \\
& \emph{Baseline-7B} & 44.71 & 81.95 & 72.73 & 0.5303 \\
& \cellcolor{mygray}\emph{Ours-7B} & \cellcolor{mygray}\textbf{53.13} & \cellcolor{mygray}\textbf{85.43} & \cellcolor{mygray}\textbf{74.66} & \cellcolor{mygray}\textbf{0.4557} \\
\arrayrulecolor{black}\bottomrule
\end{tabular}}
\vspace{-0.5cm}
\label{table:mainresults}
\end{center}
\end{table*}
\setlength{\tabcolsep}{1pt}

\subsection{Illumination Correction}
\label{sec:met-ill}  
To provide effective supervision for the feature fusion process, we design an \textbf{Illumination Correction Loss}, denoted as $\mathcal{L}_{\mathrm{IC}}$.
During training, we employ a triplet composed of a normal-light frame, an extreme-light frame, and its corresponding event frame as input. The model is supervised by our proposed $\mathcal{L}_{\mathrm{IC}}$, which guides it to distill the non-degraded semantic information from the normal-light frame into our fused feature. This loss function facilitates this process by enforcing alignment between the fused features and the normal-light features in the semantic space, thereby enhancing the model's robustness under extreme lighting conditions.
As shown in Figure~\ref{fig:framework}, we first use the frozen $\mathcal{E}_{\text{vision}}$ to extract the feature of the normal frame:
\begin{equation}
    F_{\text{original}} = \mathcal{E}_{\text{vision}}(X_{\text{original}}),
\end{equation}
which is used for calculating the mean squared error with the fused feature $F_{\text{fusion}}$ (from Section~\ref{sec:met-fea}) as the $\mathcal{L}_{\mathrm{IC}}$:
\begin{equation}
    \mathcal{L}_{\text{IC}} = \left\| F_{\text{fusion}} - F_{\text{original}} \right\|_2^2.
\end{equation}
Thus, by minimizing this discrepancy, $\mathcal{L}_{\mathrm{IC}}$ directly forces the fusion module to reconstruct features that match the semantics of a normal-light scene. Through this, the model learns to adaptively extract complementary information from events to compensate for degradation in the image. This self-supervised capability enables the model to perform illumination correction autonomously during inference, requiring only the extreme-light input and its event frames.

\subsection{Training and Inference}
\label{sec:met-train}
\noindent Based on triplet input: $X_{\text{original}}$, $X_{\text{extreme}}$, and $X_{\text{event}}$, our training is divided into two stages: 1) Self-adaptive Feature-fusion, and 2) MLLM Instruct-following Fine-tuning.

\noindent\textbf{Self-adaptive Feature-fusion.} The primary objective of the stage is to train the fusion module (Section~\ref{sec:met-fea}) to produce a refined feature representation, $F_{\text{fusion}}$, which is semantically aligned with that of a normal-lightness image. 
In this stage, we train the learnable MLPs in the fusion module to distill the non-degraded visual semantics from the normal-light image into the fused representation.
After training, the fusion module is capable of autonomously correcting for illumination degradation in the language space without the normal-light image input, effectively providing a robust and pre-aligned visual input for the subsequent fine-tuning.

\noindent\textbf{Instruct-following Fine-tuning.} The stage focuses on aligning these robust visual features after fusion with the language model for instruction-following tasks. During this phase, we fine-tune the whole event-enhanced MLLM using Low-Rank Adaptation (LoRA)~\cite{hu2022lora}. 

\noindent\textbf{Inference.} During inference, only the extreme-light image $X_{\text{extreme}}$ and event stream $X_{\text{event}}$ are available. The model fuses them into $F_{\text{fusion}}$ under the guidance of the Illumination Indicator, which is then projected and fed to the LLM for multimodal reasoning, without requiring any normal-light reference image.

\section{Experiments}
\label{sec:exp}

\noindent \textbf{Dataset and Benchmarks.} 
We design and create a large triplet (normal-light frames, extreme-light frames, and event frames) instruct-following dataset containing 2,241 samples with 10,129 question-answer pairs. For each sample, we design 17 different brightness rates as different extreme cases. We select four spatially and temporally matched event-RGB paired datasets~\cite{Gehrig21ral, li2024event, liang2024towards, liu2025ner} and utilize GPT-4o \cite{openai2024gpt4ocard} with human check to generate fine-grained language descriptions for each clip. Simultaneously, we introduce a novel benchmark featuring two tasks based on different question types: a Multiple-Choice task requiring comprehensive scene understanding with multiple correct options and an Object-Counting task requiring precise enumeration. More details are shown in supplementary material.

\begin{figure*}[t]
\centering
\includegraphics[width=0.99\linewidth]{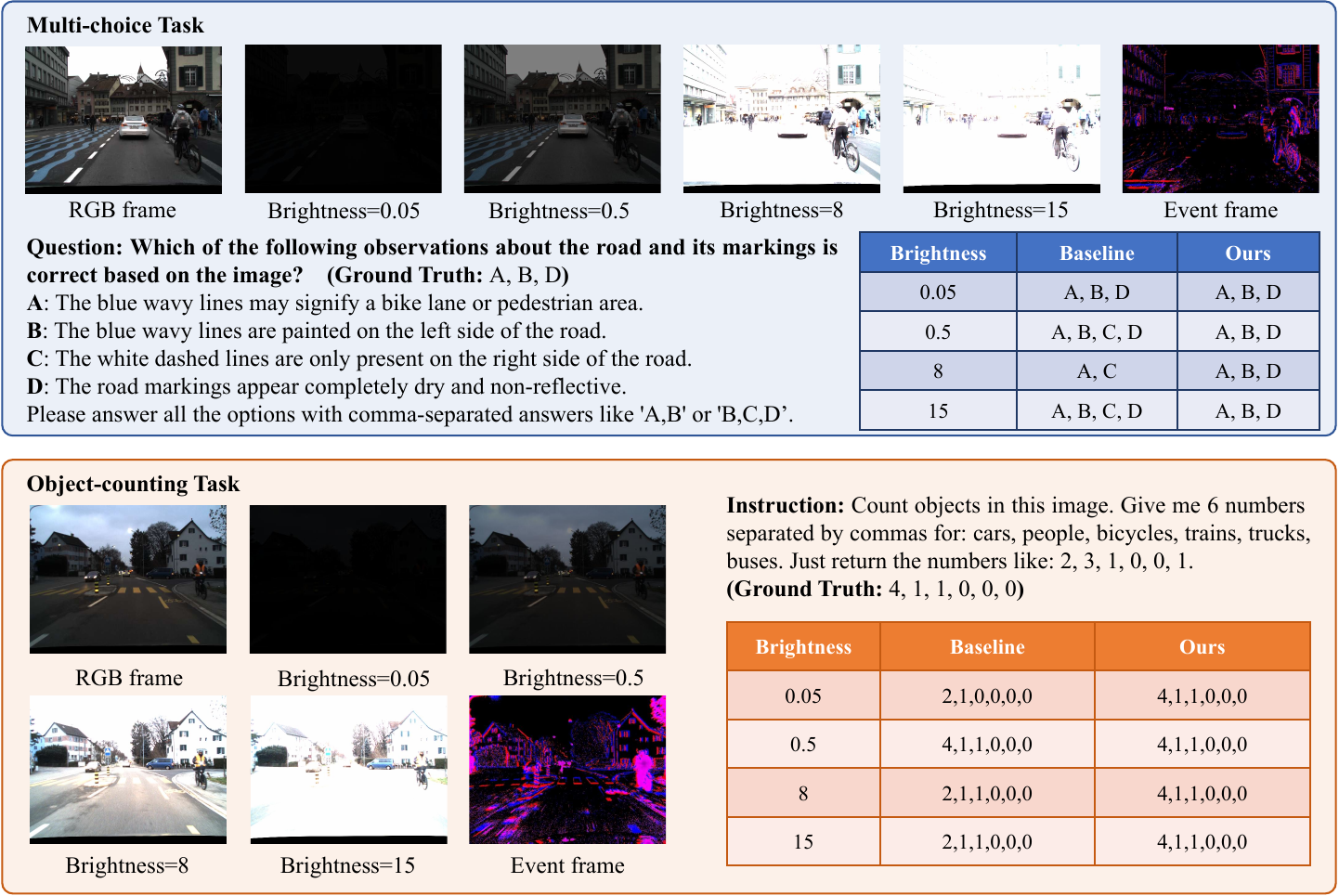}
\vspace{-0.1cm}
\caption{Qualitative comparison under various illumination levels for the multi-choice task (top) and the object-counting task (bottom). Under severe underexposure and overexposure, the baseline model frequently produces incorrect or hallucinated predictions, whereas our illumination-guided event-enhanced model remains stable and accurate across all brightness ratios. These examples illustrate the robustness of our method in extreme-light conditions.}
\vspace{-0.3cm}
\label{fig:result}
\end{figure*}

\noindent \textbf{Metrics.} 
To evaluate robustness under different lighting conditions, we systematically vary the brightness on RGB images across 17 levels, from 0.05 to 20.0 (where 1.0 represents the original). Performance is assessed using dual metrics per task: for the Multiple-Choice task, we report Accuracy (considering all options correct) and an F1 score (treating each option as a binary classification); for the Object-Counting task, we use Accuracy and Mean Absolute Error (MAE). All reported values are averaged over the 17 brightness levels, providing a comprehensive measure of both precise reasoning and error tolerance in event-visual understanding.

\noindent \textbf{Implementation Details.}
For model structure, we employ Qwen-3B and Qwen-7B~\cite{wang2024qwen2vl} as our backbones and basic baselines. We implement training on NVIDIA RTX 5090D and NVIDIA A800. For illumination correction, we employ the adam optimizer with a batch size of 4. We set the learning rate to be 0.001 and train for 30 epochs. For LoRA tuning, we train the model for one epoch.

\subsection{Comparisons with Existing Works}

We evaluate our method on our proposed benchmark, which assesses model capabilities on both multi-choice and object-counting tasks across 17 different brightness levels as shown in Table~\ref{table:mainresults}. 

To the best of our knowledge, we are the first to design a MLLM with dual input branches specifically for RGB images and event frames to address extreme illumination degradation challenges. Thus, we select three categories of existing models for comprehensive comparisons: (1) General MLLMs: LLaVA-7B~\cite{liu2023llava}, DeepSeek-VL2-Tiny~\cite{wu2024deepseekvl2mixtureofexpertsvisionlanguagemodels}, and InternVL3-8B~\cite{zhu2025internvl3exploringadvancedtraining}, which serve to evaluate inherent robustness of conventional MLLMs under extreme lighting conditions without specialized adaptation. (2) Event-specialized MLLMs: EventGPT~\cite{liu2024eventgpt} processes directly on event streams, demonstrating whether event representations alone can overcome illumination challenges that confuse conventional vision systems. (3) Illumination-aware MLLMs: Q-Instruct~\cite{zhang2024qinstruct} builds on mPLUG-Owl-2-7B fine-tuned with degraded visual conditions including extreme brightness variations, which evaluates whether existing illumination-adaptive approaches can generalize to the challenges. Meanwhile, we utilize Qwen-3B and Qwen-7B~\cite{wang2024qwen2vl} as the backbone to implement our design, and consider them as the baselines as shown in the last 4 rows of Table~\ref{table:mainresults}.

As for the quantitative results in Table~\ref{table:mainresults}, compared with existing MLLMs, our method achieves state-of-the-art performance across both evaluation tasks on our proposed benchmark. In multi-choice task, our 7B-parameter model surpasses the second-best performing model (\textit{baseline-7B}) by 8.42\% in accuracy and 3.48\% in F1 score, demonstrating though baseline models generalize comparatively well, our method still showcases substantial effectiveness in complex scene understanding under extreme lighting conditions. For object counting task, the leading \textit{Ours-7B} with 74.66\% accuracy is superior to \textit{baseline-7B} of 72.73\%, which is followed by InternVL3-7B of 72.56\% accuracy.

\begin{table*}
\caption{Ablation studies on illumination-aware adaptive feature fusion via Illumination-Correction 
(Section~\ref{sec:met-ill}) and LoRA fine-tuning (Section~\ref{sec:met-train}). Starting from the baseline (w/o both components), we evaluate the contribution of Illumination-Correction alone (w/o LoRA fine-tuning), and the full two-stage pipeline (w. Both Illumination-Correction and LoRA fine-tuning) on multi-choice and object-counting tasks. (Best results are shown in \textbf{bold}). }
\vspace{-0.6cm}
\centering
\renewcommand{\arraystretch}{1.1}
\setlength{\tabcolsep}{7pt}
\begin{center}
\resizebox{0.99\linewidth}{!}{
\begin{tabular}{llcccccc}
% \footnotesize
\toprule
\multirow{2}{*}{\textbf{Task}} & \multirow{2}{*}{\textbf{Metric}} &  \multicolumn{3}{c}{\textbf{Backbone: Qwen-3B}} & \multicolumn{3}{c}{\textbf{Backbone: Qwen-7B}} \\
& & \textbf{Baseline} & \textbf{Ill-Correction} & \cellcolor{mygray}\textbf{Ours} & \textbf{Baseline} & \textbf{Ill-Correction}& \cellcolor{mygray}\textbf{Ours} \\
\midrule
\multirow{2}{*}{Multi-Choice}& \textbf{Accuracy (\%) $\uparrow$} & 29.86 & 32.95 & \cellcolor{mygray}\textbf{34.48} & 44.71 & 51.90 & \cellcolor{mygray}\textbf{53.13}\\
& \textbf{F1 Score (\%) $\uparrow$} & 77.04 & 79.13 & \cellcolor{mygray}\textbf{80.35} & 81.95 & 84.83 & \cellcolor{mygray}\textbf{85.43}\\
\midrule
\multirow{2}{*}{Object-Counting}& \textbf{Accuracy (\%) $\uparrow$} & 69.66 & 70.48 & \cellcolor{mygray}\textbf{71.08} & 72.73 & 74.23 & \cellcolor{mygray}\textbf{74.66} \\
& \textbf{MAE $\downarrow$} & 0.6655 & 0.6194 & \cellcolor{mygray}\textbf{0.6127} & 0.5303 & 0.4645 & \cellcolor{mygray}\textbf{0.4557}\\
\bottomrule
\end{tabular}}
\vspace{-0.5cm}
\label{table:abl-ILL}
\end{center}
\end{table*}
\setlength{\tabcolsep}{1pt}

% Table: ablation on Towers
\begin{table}
\caption{Ablation studies on direct training strategies: Pre-fusion (pixel-level summation), Post-fusion (feature-level addition), and our 2 staged training method on multi-choice and object-counting tasks. We evaluate the performance comparison across Qwen-3B and Qwen-7B backbones (Best results are shown in \textbf{bold}). }
\vspace{-0.6cm}
\centering
\renewcommand{\arraystretch}{1.1}
\setlength{\tabcolsep}{5pt}
\begin{center}
\resizebox{1\linewidth}{!}{
\begin{tabular}{lcccc}
% \footnotesize
\toprule
\multirow{2}{*}{\textbf{Method}} & \multicolumn{2}{c}{\textbf{Multi-Choice QA}} & \multicolumn{2}{c}{\textbf{Object-Counting}} \\
& \textbf{Acc. (\%) $\uparrow$} & \textbf{F1 (\%) $\uparrow$} & \textbf{Acc. (\%) $\uparrow$} & \textbf{MAE $\downarrow$} \\
\midrule
\multicolumn{5}{l}{\textit{Backbone: Qwen-3B}} \\
\textbf{Pre-fusion} & 19.58 & 73.96 & 68.21 & 0.7698 \\
\textbf{Post-fusion} & 26.17 & 77.51 & 70.52 & 0.6427  \\
\cellcolor{mygray}\textbf{Ours} & \cellcolor{mygray}\textbf{34.48} & \cellcolor{mygray}\textbf{80.35} & \cellcolor{mygray}\textbf{71.08} & \cellcolor{mygray}\textbf{0.6127} \\
\midrule
\multicolumn{5}{l}{\textit{Backbone: Qwen-7B}} \\
\textbf{Pre-fusion} & 14.85 & 67.27 & 65.24 & 0.7082 \\
\textbf{Post-fusion} & 32.24 & 77.23 & 74.48 & 0.4914 \\
\cellcolor{mygray}\textbf{Ours} & \cellcolor{mygray}\textbf{53.13} & \cellcolor{mygray}\textbf{85.43} & \cellcolor{mygray}\textbf{74.66} & \cellcolor{mygray}\textbf{0.4557} \\
\bottomrule
\end{tabular}}
\vspace{-0.1cm}
\vspace{-0.4cm}
\label{table:abl-tower}
\end{center}
\end{table}
\setlength{\tabcolsep}{1pt}

% Table: ablation on Cosine Similarity
\begin{table*}
\caption{Cosine similarity comparison of vision encoder features between original and varying brightness ratios with/without illumination correction (Section~\ref{sec:met-fea}) across Qwen-3B/7B backbones. }
\vspace{-0.5cm}
\centering
\renewcommand{\arraystretch}{1.3}
\setlength{\tabcolsep}{6.5pt}
\begin{center}
\resizebox{1\linewidth}{!}{
\begin{tabular}{llcccccccc}
% \footnotesize
\toprule
\textbf{Backbone} & \textbf{Brightness Rate (Underexposure)} & \textbf{0.05} & \textbf{0.08} & \textbf{0.1} & \textbf{0.125} & \textbf{0.2} & \textbf{0.4} & \textbf{0.5} & \textbf{0.75}\\
\midrule
\multirow{2}{*}{Qwen-3B} & \textbf{w/o Illumination Correction} & 0.6597 & 0.7429 & 0.7782 & 0.8069 & 0.8593 & 0.9342 & 0.9554 & 0.9886 \\
& \cellcolor{mygray}\textbf{with Illumination Correction} & \cellcolor{mygray}0.9273 & \cellcolor{mygray}0.9368 & \cellcolor{mygray}0.9405 & \cellcolor{mygray}0.9439 & \cellcolor{mygray}0.9508 & \cellcolor{mygray}0.9619 & \cellcolor{mygray}0.9652 & \cellcolor{mygray}0.9710\\
% \midrule
\multirow{2}{*}{Qwen-7B} & \textbf{w/o Illumination Correction} & 0.7420 & 0.8011 & 0.8272 & 0.8520 & 0.8981 & 0.9551 & 0.9706 & 0.9927 \\
& \cellcolor{mygray}\textbf{with Illumination Correction} & \cellcolor{mygray}0.9185 & \cellcolor{mygray}0.9311 & \cellcolor{mygray}0.9359 & \cellcolor{mygray}0.9416 & \cellcolor{mygray}0.9508 & \cellcolor{mygray}0.9615 & \cellcolor{mygray}0.9647 & \cellcolor{mygray}0.9697 \\
\bottomrule
\textbf{w/o} & \textbf{Brightness Rate (Overexposure)}  & \textbf{2} & \textbf{3} & \textbf{5} & \textbf{7.5} & \textbf{8} & \textbf{10} & \textbf{15} & \textbf{20} \\
\midrule
\multirow{2}{*}{Qwen-3B} & \textbf{w/o Illumination Correction} & 0.9741 & 0.9556 & 0.9245 & 0.8722 & 0.8618 & 0.8187 & 0.7240 & 0.6510\\
& \cellcolor{mygray}\textbf{with Illumination Correction} & \cellcolor{mygray}0.9685 & \cellcolor{mygray}0.9637 & \cellcolor{mygray}0.9575 & \cellcolor{mygray}0.9493 & \cellcolor{mygray}0.9479 & \cellcolor{mygray}0.9404 & \cellcolor{mygray}0.9265 & \cellcolor{mygray}0.9138 \\
% \midrule
\multirow{2}{*}{Qwen-7B} & \textbf{w/o Illumination Correction} & 0.9775 & 0.9570 & 0.9184 & 0.8614 & 0.8504 & 0.8058 & 0.7089 & 0.6359 \\
& \cellcolor{mygray}\textbf{with Illumination Correction} & \cellcolor{mygray}0.9656 & \cellcolor{mygray}0.9593 & \cellcolor{mygray}0.9489 & \cellcolor{mygray}0.9381 & \cellcolor{mygray}0.9361 & \cellcolor{mygray}0.9278 & \cellcolor{mygray}0.9092 & \cellcolor{mygray}0.8939 \\
\bottomrule
\end{tabular}}
\vspace{-0.1cm}
\vspace{-0.4cm}
\label{table:abl-cs}
\end{center}
\end{table*}
\setlength{\tabcolsep}{1pt}

Moreover, scaling up the models parameter size also significantly enhances general performance in both multi-choice and object counting task (from \textit{Ours-3B} to \textit{Our-7B}, multi-choice and object counting accuracies surged from 34.48\% to 53.13\% and from 71.08\% to 74.66\%), which implies potential measures to scale-up on our approach.

As for qualitative results illustrated in Figure~\ref{fig:result}, our method demonstrates superior performance in extreme illumination conditions compared to the baseline. In the multi-choice task (top row), the model is required to identify correct statements about road markings and vehicle positions. At extremely low brightness (ratio = 0.05), both methods correctly identify that the white dashed line appears only on the right side of the road. However, under moderate darkness (ratio = 0.5) and severe overexposure (ratios = 8, 15), the baseline method exhibits hallucination by incorrectly claiming the dashed line exists on the left side or misses correct options entirely. In contrast, our event-enhanced model maintains consistent accuracy across all illumination levels by leveraging complementary information from the event camera.

For the object counting task (bottom row), the scene contains four vehicles at a distance and a prominent bicycle on the right. The baseline model only succeeds under near-normal lighting (ratio = 0.5), failing completely under extreme conditions where structural information in RGB frames is degraded. Our approach correctly enumerates objects across all listed brightness rates. This robustness stems from our illumination-aware adaptive feature fusion, which preserves stable object-level semantics even when pixel-level cues are heavily corrupted.

\subsection{Ablation Studies}
\label{sec:exp-abl}

\noindent \textbf{Architecture Design Analysis.}
We first investigate the contribution of the illumination-aware adaptive feature fusion step including Illumination Correction and the LoRA fine-tuning. In Table~\ref{table:abl-ILL}, we observe that the sequential introduction of the Illumination Correction module and LoRA fine-tuning brings incremental performance gains across all metrics for both Qwen-3B and Qwen-7B. This improvement demonstrates that each component contributes progressively to better feature alignment and reasoning capability. Notably, Qwen-7B achieves more accuracy improvements than Qwen-3B, indicating that models with larger capacity benefit more substantially from both additional modalities and sophisticated alignment methods.

\begin{figure}[t]
\centering
\includegraphics[width=1.0\linewidth]{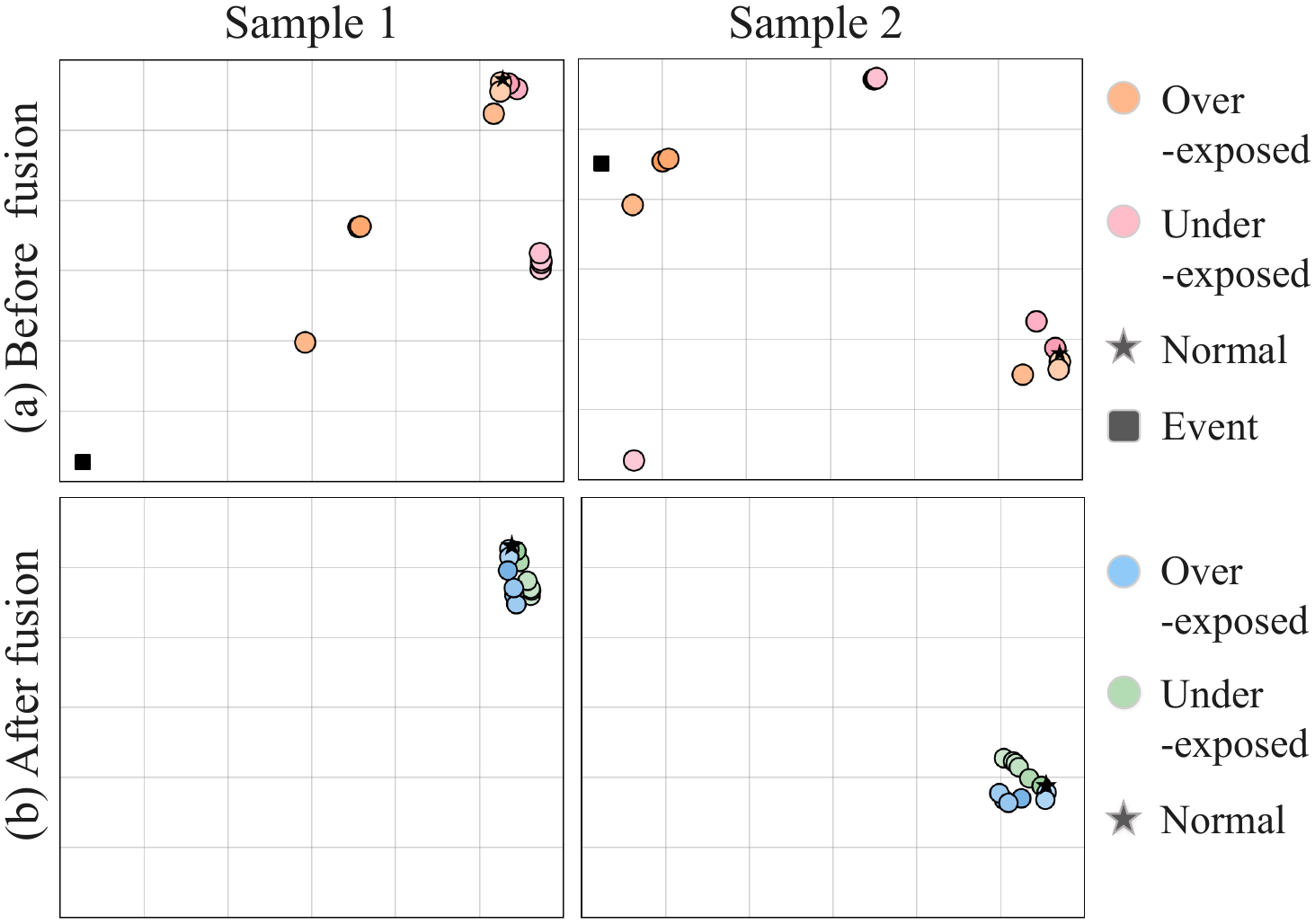}
\vspace{-0.5cm}
\caption{Visualization results of features before and after fusion with t-SNE. (a) Before fusion: event features and RGB features with different brightness ratios. (b) After fusion, fusion features under different brightness ratios cluster around the normal-light feature.
}
\vspace{-0.3cm}
\label{fig:tsne}
\end{figure}

\noindent{\textbf{Fusion Strategy Analysis}. 
We further compare our two-stage training pipeline against direct fine-tuning approaches using the same RGB-event input data. We compared with two fusion strategies including:
a) \textit{Pre-fusion}: Fusing RGB and event images pixel-wisely and feed directly into the model.
b) \textit{Post-fusion}: Features separately extracted from RGB and event images are fused via addition before feeding into the projectors and language model. The results are shown in Table~\ref{table:abl-tower}. the pre-fusion method performs poorly across all metrics. It demonstrates that the pixel-wise fusion generates a non-natural image representation, and the natural-RGB-based vision encoder fails to extract the correct features even after finetuning. The post-fusion method shows improvement over pre-fusion but remains inferior to our two-stage framework. It shows that our strategy aligning features before LoRA fine-tuning simplifies the model convergence and enhances performance.

\subsection{Feature Fusion Studies}
\label{sec:exp-abl}

This section presents an analysis of the feature alignment performance across varying illumination conditions, through the use of cosine similarity measurements as well as t-SNE visualizations. 
Table~\ref{table:abl-cs} represents a comparative analysis of cosine similarity between vision encoder features from the original images and their counterparts under different illuminations. for both Qwen-3B and Qwen-7B, the feature similarity degrades significantly under extreme brightness rates, dropping from 0.98 to approximately 0.65. In contrast, our Illumination Correction module enables the vision encoder to produce features that remain closely aligned with the original features even under extreme illumination. This demonstrates that our module effectively identifies and compensates for the missing information in RGB features by leveraging event data.

A similar trend can be observed in the t-SNE visualization in Figure \ref{fig:tsne}. In row (a), the features under varying illumination are widely dispersed, particularly under extreme ratios. It reveals that the feature shifts significantly under different illumination conditions before fusion. In contrast, row (b) shows that with our proposed fusion method, the corrected features converge and become tightly clustered around the normal-light features.

\section{Discussion}
\label{sec:disc}

\noindent \textbf{Rethinking multimodal robustness.} 
By fusing and aligning event and RGB features in the latent space, our method achieves SOTA performance and learns a shared semantic manifold that remains stable even when the RGB modality collapses. This shows that multimodal alignment can arise from principled feature-space design rather than pixel-level restoration or complex architectures. Such a representation-centric perspective suggests a promising path toward more resilient and adaptive multimodal intelligence.

\noindent \textbf{Language model affects a lot.} 
After feature alignment, we observe that both Qwen-3B and Qwen-7B produce visual features under extreme illumination that more closely match those from normal-light images. However, weaker language models show lower tolerance to residual feature variations, while stronger ones integrate multimodal cues more reliably and generate more stable responses. This highlights the importance of jointly considering feature-level alignment and the expressive capacity of the underlying language model when pursuing illumination-robust multimodal reasoning.

\noindent \textbf{Broader impacts.} 
This work investigates the use of event camera data to compensate for the visual information lost under extreme illumination by leveraging the sensor's capability to capture valuable visual cues in such conditions. This approach can be extended to other scenarios where event cameras offer advantages, like handling motion blur in RGB images. Also, the proposed paradigm can be applied to a diverse set of vision tasks, including perception~\cite{zhang2025crpq,liu2023mars3d,zhao2025equipping,zhao2025learning}, 3D reconstruction~\cite{lyu2024total,zhang2025polar,zhang2023polar}, localization~\cite{liang2024source,liang2025robust}, etc.

\noindent \textbf{Limitation.} 
While Event-MLLM significantly improves robustness under extreme illumination, it currently relies on event-camera data. It would  broaden applicability if adapting the model to scenarios where event signals are sparse or intermittently available. Moreover, expanding our multi-illumination dataset to cover more diverse real-world conditions and instruction types could further enhance generalization~\cite{chang2025far,wu2025mixture}. We regard them as promising directions toward more comprehensive all-light multimodal understanding.
\section{Conclusion}
\label{sec:conc}

We introduce Event-MLLM, a MLLM that leverages event-camera signals to overcome the inherent fragility of RGB-based perception under extreme illumination. Through an illumination-aware fusion mechanism, driven by our Illumination Indicator and Illumination Correction Loss, the model adaptively integrates complementary event and RGB cues to recover structure and semantics lost in degraded visual conditions. To support this paradigm, we curate a comprehensive multi-illumination event-instruction dataset and benchmark, enabling systematic evaluation across reasoning, localization, and fine-grained recognition. Extensive experiments demonstrate that Event-MLLM achieves state-of-the-art performance and significantly narrows the gap between extreme-lighting and normal-lighting scenarios. We believe this work establishes a foundation for integrating unconventional modalities into MLLMs and opens promising directions toward resilient, all-light multimodal intelligence.

\clearpage

\section*{Acknowledgements}

This work was supported in part by Hong Kong Research Grants Council (RGC) through the Research Impact Fund under Project R7003-21; and in part by the AI Chip Center for Emerging Smart Systems (ACCESS) sponsored by InnoHK Funding, Hong Kong, SAR, China. Meanwhile, the work has been supported by Hong Kong Research Grant Council - General Research Fund Scheme (Grant No. 17202422, 17212923, 17215025) Theme-based Research (Grant No.T45-701/22-R), and Strategic Topics Grant (Grant No.STG3/E-605/25-N). Part of the described research work is conducted in the JC STEM Lab of Robotics for Soft Materials funded by The Hong Kong Jockey Club Charities Trust.

{
    \small
    \bibliographystyle{ieeenat_fullname}
    \bibliography{main}
}


\end{document}